%% file: sample-sigconf.tex
\documentclass[sigconf]{acmart}

\usepackage[utf8]{inputenc}
\usepackage{balance}
\usepackage{amsmath}
\usepackage[mathscr]{eucal}
\usepackage[linesnumbered]{algorithm2e}
\usepackage{bm}
\usepackage{xcolor}
\usepackage{hyperref}
\usepackage{float}

\AtBeginDocument{%
  \providecommand\BibTeX{{%
    \normalfont B\kern-0.5em{\scshape i\kern-0.25em b}\kern-0.8em\TeX}}}

\acmSubmissionID{}

\begin{document}
\makeatletter
\edef\orig@output{\the\output}
\output{\setbox\@cclv\vbox{\unvbox\@cclv\vspace{0pt plus 100pt}}\orig@output}
\makeatother
\global\hbadness=10000

% \pagenumbering{arabic}
% \pagestyle{plain}
% \pagestyle{headings}

% \acmConference[DaSH@KDD]{}{August 14, 2021}{Virtual Conference}
\acmConference{-}
\acmDOI{}
\acmISBN{}

%%
%% The "title" command has an optional parameter,
%% allowing the author to define a "short title" to be used in page headers.
\title{
% \sys: 
% granular Twitter location prediction with human machine interaction
Fine-grained Geolocation Prediction of Tweets with \\ Human Machine Collaboration
% Location detection from Tweeter posts
}

\orcid{1234-5678-9012}
\author{Florina Dutt}
% \authornotemark[1]
\email{florina.design@gatech.edu}
\affiliation{%
  \institution{Georgia Institute of Technology USA}
  \city{Atlanta}
  \state{GA}
  \postcode{}
}

%%
%% The "author" command and its associated commands are used to define
%% the authors and their affiliations.
%% Of note is the shared affiliation of the first two authors, and the
%% "authornote" and "authornotemark" commands
%% used to denote shared contribution to the research.
\author{Subhajit Das}
\email{das@gatech.edu}
\affiliation{%
  \institution{Georgia Institute of Technology USA}
  \city{Atlanta}
  \state{GA}
%   \postcode{}
}

\newcommand{\sd}[1]{\textcolor{blue}{[S: #1]}}
\newcommand{\todo}[1]{\textcolor{gray}{[#1]}}
\newcommand{\fd}[1]{\textcolor{orange}{[F: #1]}}

\newcommand{\eg}{\textit{e.g.}}
\newcommand{\ie}{\textit{i.e.}}
\newcommand{\etal}{\textit{et al.}}
\newcommand{\sys}{LocDet}
\newcommand{\user}{Trace~}
\newcommand{\userB}{Shana}
\newcommand{\cityA}{New York City~}
\newcommand{\cityB}{Atlanta~}

\renewcommand{\algorithmautorefname}{Algorithm}
\renewcommand{\equationautorefname}{Equation}

\definecolor{light-gray}{gray}{0.95}
\newcommand{\code}[1]{\colorbox{light-gray}{\texttt{#1}}}

%%
%% By default, the full list of authors will be used in the page
%% headers. Often, this list is too long, and will overlap
%% other information printed in the page headers. This command allows
%% the author to define a more concise list
%% of authors' names for this purpose.
% \renewcommand{\shortauthors}{Das and Xu, et al.}

\input{content/abstract.tex}

\maketitle
% \teaser{
%  \includegraphics[width=\textwidth]{img/teaser_1.PNG}
%  \caption{Interface of \sys showing the two main views}
% }
% \begin{figure*}
% \includegraphics[width=1\textwidth,height=8.1cm]{img/teaser_3.PNG}
%   \caption{Main views of \sys: 
%   A. Heatmap matrices. 
%   B. Topic scatterplot and bar chart view showing top keywords per topics.
%   C. Word-cloud showing top keywords.
%   D. Stacked bar chart showing number of correct sentiment predictions with time.
%   E. Confusion matrix.
%   F. Model overview showing sentiment classifiers constructed per iteration.
%   G. Search bar to filter data.
%   H. Text view to add/remove/replace and specify text length to training data.
%   I. Mouse over heatmap to see Text View. \looseness = -2
%   }
%   \label{fig:teaser}
% \end{figure*}

\input{content/intro.tex}

\input{content/related_works.tex}
\input{content/system_technique.tex}

\input{content/conclusion.tex}

\bibliographystyle{ACM-Reference-Format}
\bibliography{sample-base}
\balance

%%
%% If your work has an appendix, this is the place to put it.
\newpage

\appendix

\end{document}

%% file: content/abstract.tex
%%
%% The abstract is a short summary of the work to be presented in the
%% article.
\begin{abstract}

Twitter is a useful resource to analyze peoples' opinions on various topics. Often these topics are correlated or associated with locations from where these Tweet posts are made. For example, restaurant owners may need to know where their target customers eat with respect to the sentiment of the posts made related to food, policy planners may need to analyze citizens' opinion on relevant issues such as crime, safety, congestion, etc. with respect to specific parts of the city, or county or state. As promising as this is, less than $1\%$ of the crawled Tweet posts come with geolocation tags. That makes accurate prediction of Tweet posts for the non geo-tagged tweets very critical to analyze data in various domains. In this research, we utilized millions of Twitter posts and end-users domain expertise to build a set of deep neural network models using natural language processing (NLP) techniques, that predicts the geolocation of non geo-tagged Tweet posts at various level of granularities such as neighborhood, zipcode, and longitude with latitudes. With multiple neural architecture experiments, and a collaborative human-machine workflow design, our ongoing work on geolocation detection shows promising results that empower end users to correlate relationship between variables of choice with the location information.\looseness = -2

\end{abstract}

%  By leveraging prototype learning we enable user-friendly explanation. 

%%
%% The code below is generated by the tool at http://dl.acm.org/ccs.cfm.
%% Please copy and paste the code instead of the example below.

%%

% \ccsdesc[500]{Computer systems organization~Embedded systems}
% \ccsdesc[300]{Computer systems organization~Redundancy}
% \ccsdesc{Computer systems organization~Robotics}
% \ccsdesc[100]{Networks~Network reliability}

%% Keywords. The author(s) should pick words that accurately describe
%% the work being presented. Separate the keywords with commas.
\keywords{Classification, Regression, Sequence data, Twitter API, Transformers, Natural language processing, Human-centered Machine Learning}

%% A "teaser" image appears between the author and affiliation
%% information and the body of the document, and typically spans the
%% page.
% \begin{teaserfigure}
%   \includegraphics[width=\textwidth]{sampleteaser}
%   \caption{Seattle Mariners at Spring Training, 2010.}
%   \Description{Enjoying the baseball game from the third-base
%   seats. Ichiro Suzuki preparing to bat.}
%   \label{fig:teaser}
% \end{teaserfigure}

%%
%% This command processes the author and affiliation and title
%% information and builds the first part of the formatted document.

% \outline{subject to change: our technique validated on text, time-series, and image classification tasks.}

%\delete{authenticity (or faithfulness to the ordiginal model), interpretability through additional constraints - sparsity, non-negativity and therefore additive combination of prototypes, row in W adding to 1, prototype vectors constrained to be the latent space representation of real data or segments of real data..., generalized application, model and data agnostic}

%% file: content/intro.tex
\section{Introduction}
\label{sec:intro}

\begin{figure}[H]
        \includegraphics[width=3.3in]{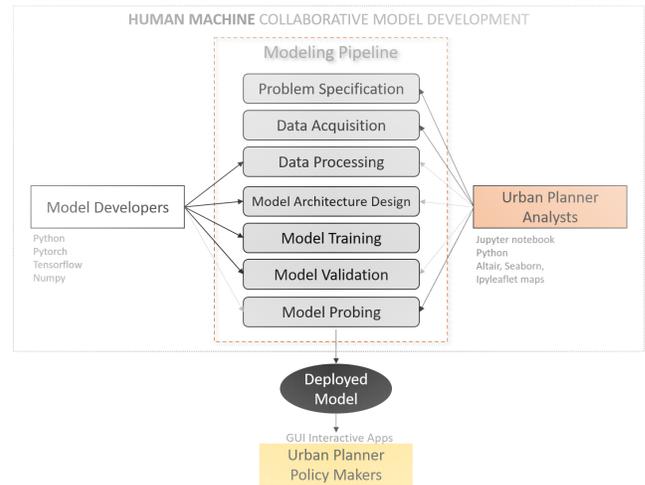}
        \caption{Workflow adopted in this collaborative model development between model developers and domain experts. \looseness = -2}
        \label{fig:diag_flow}
        \vspace{-0.3cm}
    \end{figure}

% PROBLEM DEFINITION
Machine learning (ML) has been effectively used in many real-world data analytic problem scenarios (e.g., in marketing, finance, healthcare, etc. \cite{appml}). In many domains, analysts' intend to make sense of peoples' opinion to make informed business decisions. For example, in location based advertisements user reviews are analysed with location, in citizen opinion mining tasks \cite{su11154016, useparticip} Twitter posts are used to analyse topics and entities with geolocation \cite{topicmodel_tweet}. 
In the literature and in real world analysis problems, Twitter posts (often crawled using Pythons' Tweepy API \cite{tweepy}) are a good data source for conducting such analysis. However, geolocated Tweets are very rare (less than 1\%) \cite{Miyazaki2018TwitterGU} which makes it impossible to utilise the entire data corpus for analysis. In this paper, we present our on-going work to predict Twitter posts' geolocation at a granular level such as neighborhood, zipcode, and longitude with latitude values. While geolocating Twitter posts has been researched in the past \cite{survey_zheng, Xu2020SurveyOU, city_geo}, granular location prediction \cite{chong_locdet_fine, fine_grained_user} is rare and under-explored, which is often critical for many domains.
In this work, we collaborate with urban planning analysts to not only assess the success of our modeling task by an iterative modeling pipeline but also incorporate their valuable domain expertise in making modeling decisions to refine model performance (see Figure~\ref{fig:diag_flow}). End-users (urban planners) explore and analyse the model output results using interactive visualizations in a Jupyter notebook environment. In the process they spot check results (of location prediction) and provide domain-specific critical feedback, that is programmatically implemented to: (1) clean and transform data, (2) re-design the models' architecture, and (3) extract guidelines to inform processes to further collect training data. \looseness = -2

% WHAT OTHERS HAVE DONE ... WHAT WE DID

We present a set of best performing models in addition to an in-depth summary of various model experimentation results. We discuss the successes and failures of various modeling approaches for our problem, some of which were partly inspired from the literature (see section \ref{sec:rel}), while many other algorithmic approaches were driven by feedback from domain experts (urban planners) as they probed the trained model in a Jupyter notebook environment with interactive visualizations.
%EXPERIMENTS AND ON GOING WORK WITH URBAN PLANNERS
In summary, we contribute: (1) A novel text modeling approach informed with feedback from domain expert users to predict geolocation information for non geo-tagged Tweet posts at zipcode, neighborhood, and longitude with latitude granularity. (2) Results including successes and failures from an on-going work with urban planner analysts of a real world ML problem where geolocated Tweets are needed to assess citizens' opinion on various topics critical to craft new urban policies. \looseness = -2

% \begin{itemize}
    % \noindent \textbf{ - } A novel text modeling approach informed with feedback from domain expert users to predict geo-location for non geo-located Tweet posts at zipcode, neighborhood, and longitude with latitude granularity. \looseness = -2
    
    % \noindent \textbf{ - } Results including successes and failures from an on-going work with urban planners of a real world ML problem where geolocated Tweets are needed to assess citizens' opinion on urban policy related topics. \looseness =-2

% 	A. Comparison of peoples' sentiment across cities with respect to topic of discussion and land use types eo empower urban planners satisfy data analytic goals, and B. Comparison of businesses across multiple cities using Yelp business review data proving the generalizability of the approach .

% \end{itemize}

%% file: content/related_works.tex
\section{Related Work} \label{sec:rel}

Predicting geolocations from Tweet posts has been investigated extensively. Luke et al. built an interactive system to predict geolocation for Tweets at the level of a city \cite{city_geo}. Yasuhide et al. showed integrating text, meta data and user network to build a neural network with attention mechanism to predict location information \cite{miura-etal-2017-unifying}. These surveys further provide a comprehensive overview of location detection research from Twitter posts \cite{survey_zheng, Xu2020SurveyOU}. Miyazaki et al. employed graph embedding on a knowledge graph to build an end to end geolocation classifier \cite{Miyazaki2018TwitterGU}. Tien et al. discussed MENET, a multi entry neural network architecture combining deep neural network and multi view learning that outperformed three publicly available datasets on city level prediction (metric used was Acc@161) \cite{Do2017MultiviewDL}. Chieh et al. explored multi-headed self attention model for text representation to predict location at the level of a city \cite{ross_locdet}. While these works were effective, they were not granular location prediction methods to predict a Tweets location. Aligned to this, Wen-haw et al. used a hidden markov model to geolocate tweets to a closest venue such as restaurants, shops, and others \cite{chong_locdet_fine}.  Matsuno et al. proposed a technique to predict a Twitter users' predicted location visit to a block level \cite{block_matsuno}.
Inspired from these works, we further geolocate tweets to a zipcode, neighborhood or longitude and latitude.
\looseness = -2

Li et al. discussed the application of fine-grained geolocation based on extraction of location related entities
from non-geotagged Tweet posts \cite{fine_grained_user}.
Iso et al. deployed a convolutional mixture density network to predict Tweets' geolocation along with ambiguity in prediction based on a probability distribution. Their best model predicted a Tweet with an mean error of 91 miles \cite{dens_est}.  Zola et al. estimated a users' home location using gaussian mixture models and the DBSCAN clustering approach \cite{DBscan_twit}.
Li et al. showed that their model outperformed others when trained using diverse data sources including textual content, user location, place labels and bounding box of place \cite{covid_geoloc}. Mircea et al. prototyped a classification, geolocation and interactive visualization of COVID-19 tweets. Their dashboard helped users to geolocate Tweets by fine tuning a novel L2 classification layer \cite{mircea-2020-real}.
These works inspired us to: (1) continue research in this direction, and (2) fill the missing gap of more direct human machine collaboration in model development to predict fine-grained geolocation information from social media data. \looseness =-2

%% file: content/system_technique.tex
\section{Technique}

\subsection{Problem specification}
\vspace{0.1cm}
\noindent \textbf{Dataset:}
The data was crawled from Twitter using Pythons' Tweepy library \cite{tweepy} between \textit{Jan 2018} to \textit{Dec 2020}, within 100 mile radius of Atlanta, Georgia, USA. In total, there were 10 million Tweets, out of which only 90000 Tweets were geo-tagged. We filtered the data using keywords specific to urban planning domain (provided by the urban planners), to sample an useful set of 2 million Tweet posts. This also helped discard a large set of redundant Tweet posts related to various kinds of product/service advertisement including jobs, restaurants, product promotions, and others. We set 80000 data samples for training set, 10000 for validation set, and the rest of the data (2 million - 90000 samples) for the hold out/test set. In addition, we also validate our models' performance on $200000$ Tweets from the US Global dataset \cite{tweetData} comparing geolocation prediction over Tweets from multiple cities in USA (see section \ref{sec:modelcompare} and Figure~\ref{fig:mapview}). Each Tweet from both datasets contained attributes such as \textit{Tweet-text}, \textit{Created-at}, \textit{Num-Hashtags}, \textit{Username}, \textit{Retweet-count}, and others. For geo-tagged tweets, the data also contained the attributes \textit{Longitude} and \textit{Latitude}. \looseness = -2

\vspace{0.1cm}
\noindent \textbf{Pre-processing:}
Prior to inputting the data to the modeling pipeline, the data is pre-processed and transformed in ways which improves the performance of the model.
% Each Twitter post contains the text data column, in addition to other meta data columns such as \textit{created At}, \textit{hashtags}, \textit{number of retweets}, \textit{username}, and others \cite{city_geo, Miyazaki2018TwitterGU}
Each Tweets text can be expressed as a list of words $W$.
We pre-process $W$ using standard natural language processing methods including stemming, lemmatization, removal of stop words and special characters, and limiting text length (user-specified hyperparameter) \cite{PLUNZ2019235, inmacs_dash}. 
Further pre-processing is applied as users interact and explore the data by filtering relevant Tweets, specifying more relevant class labels, adjusting hyperparameters, probing models through visualizations as seen in Figure~\ref{fig:diag_flow}. \looseness = -2

\vspace{0.1cm}
\noindent \textbf{Task:}
With the crawled data, urban planners intended to analyse peoples' sentiment, as expressed in their opinions through Twitter posts. In this human machine collaboration, they strive to design a deep learning model that can predict geolocation of non geo-tagged Tweets with highest accuracy. More importantly, their data analytic workflow required the geolocation prediction to be fine-grained, at-least 30 miles or less in prediction error. \looseness = -2

\subsection{Model design and selection}
In the following we describe a series of model architecture and hyperparameter settings (see Table~\ref{table:model_results}), each designed for the specified task (classification or regression) by the user (urban planners). In general, our approach was to design
an end-to-end ML modeling pipeline that given an input text sequence (e.g., a Tweet post) predicts the geolocation of the post such as zipcode, neighborhood or longitude with latitude as specified by the user. \looseness = -2

\vspace{0.1cm}
\noindent \textbf{Longitude and Latitude prediction:}
% PURE CNN TEXT MODEL FOR REGRESSION WITH PAIRWISE DISTANCE
Initially, we posed the problem as a regression task, as we were motivated to predict longitude and latitude. We utilised a 1-dimensional convolution based CNN for sentence classification model ($S$) as described by Kim et al. \cite{kimconvolutional}.
We trained $S$ with a wide range of kernels of size ${2,4,8,16,32,64}$ and with $128$ channels each. We applied max pooling after each convolutional layer. In addition, we added $2$ fully connected layer. Other hyperparameters included learning rate (varied between $0.001$ to $0.09$ and batch size, varied between $128$ to $2048$). Finally, with $70$ epochs, $0.005$ learning rate, and batch size of $256$ we observed better performing results.
In this model architecture, we modified the output layer to contain two nodes predicting longitude and latitude of a given input Tweet post. We utilized the pairwise distance loss function \cite{pairwise_loss} to assess the performance of this model ($ \sqrt{\sum_{0}^{n} (p_{i} - q_{i})^{2}}$).
While this model performed relatively well on the training set (acc@30miles: 47.28\%, average distance between predicted location < 38.23 miles, loss: 0.45), the performance was very poor on the validation set (acc@30miles: 12.11\%, average distance between predicted location <= 250 miles, loss: 9.453). 
We asked domain experts to explore word/token distribution present in Tweets with respect to longitude and latitude
% the model weights and its latent representation 
through visualizations in the Jupyter notebook, to help understand if there was something peculiar about the input data and the target label distribution.
% the model design or the data input shape. 
Upon probing, they found that there was a lack of any correlation between the urban planning relevant tokens (including bi-grams and tri-grams) (e.g., "green park", "biking along the river", "stuck in traffic") with the longitude and latitude information.  \looseness = -2

% COMBO MODEL
To overcome the inherent noise in the data, we were inclined to model character-level nuances to predict fine-grained location. We utilized the model architecture of a character-level CNN from the work of McKenzie et al. \cite{charcnn_geo} and others \cite{liu_charcnn}. However, we did not want to lose the information learned through the word level CNN model as described above. In this, we were motivated by the approach shown by Ross et al. \cite{ross_locdet} in predicting cities. In their model architecture they concatenated the input from two parallely trained models to make predictions. We first trained the character CNN model $C$ and retrieved the final fully connected layers' output as $e$. Next, we trained the sentence level modified CNN model $S$ (as described above); however, before the final layer making the prediction, we concatenated the output from the fully connected layer with $e$. This model architecture design helped to feed character level nuances into the modeling problem. 
% Following the same we encoded the n grams as a vector and merged with the penultimate layer to concatanten before making the prediction. 
This approach (CCH) improved the validation sets' performance  (acc@30miles: 28.15\%, average distance between predicted location <= 100 miles, loss: 2.322).
Furthermore, we replaced the model $S$ with a word level LSTM model \cite{zhang-lstm-sentence} modified with an attention layer $A$, inspired by the work of Vaswani et al. \cite{NIPS2017_3f5ee243} in the above pipeline. We utilized attention mechanism to calculate a soft alignment score between the hidden states for each token and the final hidden state representation in the LSTM model. $A$ first computes weights for each sentence in LSTM output and finally computes the updated hidden state.
% mechanism by first computing weights for each of the sequence present in lstm_output and and then finally computing the                 new hidden state.
With this model (CCH-A, $T$, containing $C$ and $A$) we noticed a slight improvement in performance over the validation set (acc@30miles: 36.09\%, average distance between predicted location <= 85.43 miles, loss: 0.322). 
\looseness = -2
% worked as the validation set performance improved slightly (acc: 39.43\%, loss 2.32).

\vspace{0.1cm}
\noindent \textbf{Zipcode prediction:}
% \noindent \textbf{Label acquisition:}
% The previous model excersize failed to give satisfactory results for the intended fine grained geolocation prediction.
% While these datasets contained \textit{Longitude} and \textit{Latitude} as target variables, our first iteration of model architecture design 
Models trained to predict \textit{Longitude} and \textit{Latitude} showed less than satisfactory results. Based on feedback from urban planners, we decided to predict zipcodes. To retrieve zipcodes for the geo-tagged data, we utilized Arcgis' reverse geocoding API for Python \cite{arcgiscode}, which given an input tuple of longitude, latitude pairs, returned a JSON object containing the address of the location. This JSON object included street name, county, neighborhood name, and the zipcode. The data 
% contained Tweets from the state of Georgia, USA. That 
contained 727 zipcodes from the state of Georgia, USA, as target labels in the input data. Now our task was to build a classifier instead of a regressor. \looseness =-2

Our next model architecture 
% was an end-to-end ML modeling pipeline that given an input text sequence (e.g., a Tweet post) predicts the geolocation of the post. It 
is based on a modified transformer model (with multi-headed attention layer) \cite{NIPS2017_3f5ee243} to predict the zipcode of a Tweet (categorical target variable). This model (MH) $M$ is different from a conventional transformer model in the following ways: (1) we discarded the decoder from the transformer model, (2) we removed residual connections, layer normalization, and the layer masking (as its' a classification task, not a language modeling problem), and (3) we employed a multi-headed attention with position-wise feedforward encodings. These changes are inspired by the implementation referred here \cite{model_gits}. Since this was a classification task to predict zipcodes the final layer incorporated a cross entropy loss with soft-max activation function ($ -\sum_{0}^{n} y_{i} log(J(f_{0}(x_{i})))$, J is softmax function).  
% pure multi headed results
When trained, this model $M$ performed very highly on the training set (acc: 83.32\%, loss: 0.12) but poorly on the validation set (acc: 22.23\%, loss: 1.65). We specified batch-size: 512, learning-rate: 0.004, number of encoders: 3, embedding dimension: 1024, division-factor: 128, number-epochs: 50 as hyperparameters to this model, and optimized using the Adam optimizer \cite{kingma2014method_adam, various_optim}. Aiming to reduce overfitting, we specified a drop out rate of 0.3, reduced batch-size to 64 and utilised a learning rate decay function. However, this did not improve the generalizability of the model as per expectation.  \looseness = -2

%with username dict and validation set embellishement
We brainstormed with the urban planners as we reviewed the text content of the Tweets and the set of users who were the authors of the Tweets in our data corpus. Upon investigation, we found that there were many users who were posting content more frequently than others. We also found that they shared similar content especially within a close time range (say within a weeks time) from a nearby geolocation (e.g., within 10 miles of each post). Based on this insight, we constructed a username dictionary object $U$ (from the training set) that stored per user $u_i$: (1) posted Tweet Ids $ w_1, w_2, w_3 .... $, Tweet creation date/time $tm_1, tm_2, tm_3 .... $, and their geolocation information $l_1, l_2, l_3$. Next, only for validation set, during prediction we query for the username from $U$. If the username is in the dictionary we retrieve all the tweets with the tweet ids $w_1, w_2, w_3 .... $, for that user. Next, based on the creation date of the Tweet in the validation set, we sample a subset of tweets $f$ from $U$ (specified by a given time range by users which is a hyperparameter, e.g., find Tweets within 2 days of post). Finally, we sample the first $k$ (an hyperparameter) tokens of each tweet in $f$ and concatenate with the queried Tweets text content. The intuition of this approach directly stemmed from the observations made by the urban planners upon analysing the training corpus. Their insight was that users usually post on similar topics/content within a close time range from close-by geolocations. For example, posts about review of a restaurant within a weeks time tend to contain similar text content. Furthermore Wen-haw et al. \cite{chong_locdet_fine} followed a similar approach in predicting fine-grained geolocation for non geo-tagged Tweets. However, unlike their approach, we do not sample users' Tweet post based on their visit to a same venue such as restaurant, shops, museum etc. Using this approach (MH-U), we observed an increase in validation set performance of zipcode prediction (acc: 47.30\%, loss: 0.72). \looseness = -2

% multi headed with clustering
Next, motivated to improve the validation sets' accuracy, we further investigated other approaches that can be adopted using text content from usernames in the corpus. In this approach, instead of building a username dictionary object, we first prepared a embedding matrix of all the Tweets in the training corpus $e$. To generate the embeddings per Tweet, we utilized Tensorflows' universal sentence encoder \cite{univ_sent_enc}. Next we clustered the embedding matrix using the KMeans clustering model \cite{kmeans} with a user-specified number of clusters, which is a hyperparameter. As we train the model, it first queries the cluster membership of the username of a given Tweet. Next from $e$ it samples $nc$ (a hyperparameter specified by user) closest matching tweets (defined by cosine similarity metric) from the found cluster as $g$. The model consumes $g$ by passing it through its' encoder layer, and then concatenating the results of the encoder layer with the encoded representation of the input text as $o$. Finally $o$ is passed through a fully connected layer to make zipcode prediction. We were happy to see a significant increase in the models validation set performance (acc: 59.43\%, loss: 0.18) with this approach (MH-C). Upon inspection by the urban planners, we observed that the zipcode distribution was skewed in the validation set. With stratified sampling on the validation set (MH-C-S) and tuned hyperparameters (see Table \ref{table:model_results}) the performance further increased to acc: 67.30\%, loss: 0.09. \looseness = -2

\begin{table}[t]
  \caption{Various model performance results on our Twitter data.}
  \label{table:model_results}
%   \scriptsize%
	\centering%
    \begin{tabular} {c c c c c c}
    \hline
    Model & Acc. (valid) & Loss & Key-Params & Target  \\
    \hline
    1D-CNN & 12.11\%  & 9.453 & kernels: (2 - 64) & Long,  \\
    & & & fc layers: 2 & Lat  \\
    & & & l-rate: 0.04 & &  \\
    \hline
    
    CCH & 28.15\%  & 2.322 & kernels: (4 - 32) &  Long,  \\
    & & & fc layers: 1 & Lat  \\
    & & & l-rate: 0.09 & & \\
    \hline
    
    CCH-A & 36.09\%  & 0.322 & kernels: (2 - 64) &  Long,  \\
    & & & fc layers: 3 & Lat \\
    & & & l-rate: 0.005 & & \\
    \hline
    
    % zipcodes
    MH & 22.23 \%  & 1.65 & encoders: 3 &  Zipcode  \\
    & & & emb-dim: 3 & &  \\
    & & & div-fac: 128 & &  \\
    & & & drop-out: 0.3 & &  \\
    \hline
    
    MH-U & 47.30\%  & 0.72 & encoders: 3 &  Zipcode  \\
    & & & emb-dim: 3 & &  \\
    & & & div-fac: 128 & &  \\
    & & & drop-out: 0.3 & &  \\
    & & & k-token: 16 & &  \\
    & & & time: 72 hours & &  \\
    \hline
    
    MH-C & 59.43\%  & 0.182 & encoders: 3 &  Zipcode  \\
    & & & emb-dim: 3 & &  \\
    & & & div-fac: 128 & &  \\
    & & & drop-out: 0.3 & &  \\
    & & & num-clus: 4 & &  \\
    & & & samples: 20 & &  \\
    \hline
    
    MH-C-S & 67.30\%  & 0.091 & encoders: 4 &  Zipcode  \\
    & & & emb-dim: 3 & &  \\
    & & & div-fac: 256 & &  \\
    & & & drop-out: 0.33 & &  \\
    & & & num-clus: 9 & &  \\
    & & & samples: 35 & &  \\
    \hline
    
    %neighborhood
    MH-N-E & 71.32\%  & 0.003 & encoders: 3 &  Neighbor  \\
    & & & emb-dim: 3 & -hood  \\
    & & & div-fac: 128 & &  \\
    & & & drop-out: 0.1 & &  \\
    & & & sam-fac: 0.65 & &  \\
    \hline
    
    MH-N & 58.34\%  & 0.194 & encoders: 3 &  Neighbor \\
    & & & emb-dim: 3 & -hood   \\
    & & & div-fac: 128 & &  \\
    & & & drop-out: 0.1 & &  \\
    \hline
    \end{tabular}

\end{table}

\begin{table}[t]
  \caption{Various model performance results on US Global Twitter data \cite{tweetData}.}
  \label{table:model_results}
%   \scriptsize%
	\centering%
    \begin{tabular} {c c c c c c}
    \hline
    Model & Acc. (valid) & Loss & Key-Params & Target  \\
    \hline

    CCH-A & 39.59\%  & 0.632 & kernels: (4 - 16) &  Long,  \\
    & & & fc layers: 2 & Lat \\
    & & & l-rate: 0.046 & & \\
    \hline

    MH-C-S & 65.03\%  & 0.09 & encoders: 3 &  Zipcode  \\
    & & & emb-dim: 3 & &  \\
    & & & div-fac: 256 & &  \\
    & & & drop-out: 0.33 & &  \\
    & & & num-clus: 9 & &  \\
    & & & samples: 35 & &  \\
    \hline
    
    %neighborhood
    MH-N-E & 72.84\%  & 0.01 & encoders: 3 &  Neighbor  \\
    & & & emb-dim: 3 & -hood  \\
    & & & div-fac: 128 & &  \\
    & & & drop-out: 0.1 & &  \\
    & & & sam-fac: 0.65 & &  \\
    \hline

    \end{tabular}

\end{table}

% learning algorithm (e.g., \textit{CNN-Text model}~\cite{kimconvolutional}) and a set of hyperparameters (e.g. learning-rate). 

\vspace{0.1cm}
\noindent \textbf{Neighborhood prediction:}
While we were successful to train a model with significantly better (compared to longitude level) prediction results with respect to zipcodes, urban planners required the model to also predict neighborhood of a given tweet for their analysis process. Their rational was that neighborhoods are much more comprehensive and relatable as compared to zipcodes to make sense of analysis results in their domain. In addition, typically the sizes of the zipcodes show a high variance, some zipcodes can be very small in area, while many others cover a large geographic region. 
This variance in the zipcode size further creates ambiguity in the prediction. Neighborhood level sentiment analysis of Tweets allows urban planners to compare these sentiment scores from peoples' opinion with other factors such as built environment characteristics such as street lights, side walk depth, and others.
% Finally there are way too many zipcodes per state (over 700 for the state of Georgia), compared to the number of neighborhoods.
As mentioned before, the JSON object retrieved as address from the ArcGIS Api contained neighborhood data. However, most of the data samples in the JSON object had "null" values for the neighborhood key.
% We could have used the neighborhood key from the Arcgis API for python, but many of the data samples had null value. 
To solve this problem, with the help of the urban planners we were able to use a neighborhood shape file \cite{atl_stats_neigh} for the city of Atlanta metropolitan area in Georgia, USA. For each geo-tagged Tweet, using ArcGIS' python library Arcpy, we spatially joined the neighborhood shape file to fetch the neighborhood label per Tweet. In total there were $102$ neighborhoods in our data corpus.

To predict the neighborhood, we utilized a similar pipeline as we did to predict zipcodes, that included: (1) training a multi-headed attention model, and (2) creation of a user dictionary object to add text content to Tweets in the validation set. However, in the case of the neighborhood label prediction task, we observed a class imbalance. We thus over-sampled the minority class to balance the data. However, this significantly increased the training time and the memory requirement in our pipeline. We thus introduced a \textit{sample-factor} hyperparameter, that ranged between $0$ (means no oversampling)  to $1$ (means complete oversampling with each class balanced). We observed that the value $0.65$ worked well for our use case and task. Next, we utilized the glove embeddings instead of the universal sentence encoders to create the embedding matrix $e$. Glove embedding was not only faster to compute but also resulted in improved performance. With this model architecture design we were able to significantly improve validation sets' performance (acc: 71.32\%, loss:0.003).\looseness = -2

\subsection{Generalizing models}
\label{sec:modelcompare}
\begin{figure}[H]
        \includegraphics[width=3.3in]{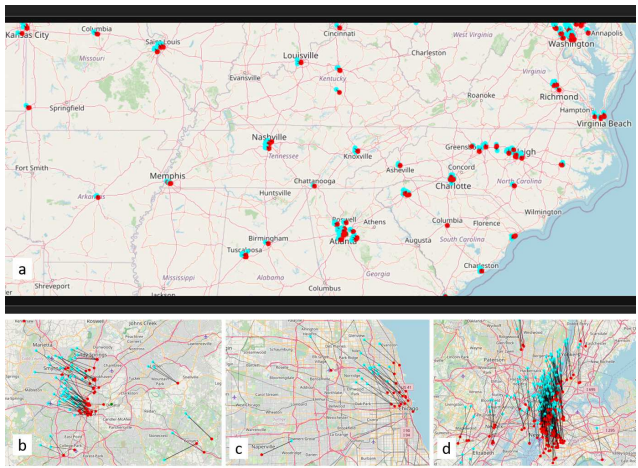}
        \caption{Visualization in Jupyter notebook of predicted geolocations (shown in blue dot) compared to ground truth (shown in red dot) for: a. USA, b. Atlanta, c. Chicago, d. New York \looseness = -2}
        \label{fig:mapview}
        % \vspace{0.3cm}
    \end{figure}

From the best set of models that we trained, we also tested their performance on the US Global Tweets dataset \cite{tweetData} (see Figure~\ref{fig:mapview}). Of the 200000 Tweets, we set 150000, 20000, and 30000 for training, validation, and test/hold out set respectively. We used the CCH-A model to predict longitude and latitudes and observed satisfactory performance (acc@30miles: 32.12\%, average distance between predicted location <= 109.66 miles, loss: 0.853). We tuned the hyperparameters to observe a little improvment in performance (acc@30miles: 39.59\%, average distance between predicted location <= 81.01 miles, loss: 0.632).
Next we trained the MH-C-S model to predict zipcodes (in total 5000 zipcodes). Upon investigating further, we remove zipcodes with less than 10 Tweet samples. The final dataset had 170000 samples (130k training, 20k validation, and 20k test set) with 900 zipcodes. When trained we observed acc: 65.03\%, loss: 0.09 on the validation set.
Finally we trained the model MH-N-E on this data and observed acc: 68.66\%, loss: 0.11 on the validation set. Upon changing the oversampling hyperparameter and discarding low frequency neighborhood Tweet samples we observed a bit higher performance (acc: 72.84\%, loss: 0.01) on the validation set. With this excersize, we ensured that the model designs chosen for this analytic pipeline were generalizable to other datasets as well. \looseness = -2

\section{Experiments}
In this section, we describe a set of ablation studies to learn the impact and role of various components in our modeling pipeline. The data used in these studies are same as the one on which we trained models (i.e, Twitter posts within 100 mile radius of Atlanta between 2018 to 2020). \looseness = -2

\vspace{0.1cm}
\noindent \textbf{User data dictionary:}
We were inquisitive to understand the relevance of the user data dictionary in modeling the location prediction problem. To that end, we ran an ablation study in which we varied the parameter that controlled the number of Tweets to sample from the user object. It ranged from $0$ (meaning no Tweet will be sampled) to $1$ (meaning all Tweets from that user will be sampled). When the parameter was set to $0$ we found very poor validation set accuracy (acc: $22.2\%$), while with the parameter set to $0.5$ the validation set accuracy maximized (acc: $45.22\%$). Upon further setting the parameter to $1$, we found mild reduction in performance (acc: $41\%$). Thus we understand that there is a sweet spot in which relevant text from the same user when added to the Tweet leads to increasing the performance. Adding every possible Tweets' content (of the same user) leads to adding noise, which causes drop in the models' performance.
We performed a similar study with the parameter that controlled the time range within which relevant Tweet from the same user will be sampled. We ranged it between $0$ (means no Tweet will be sampled) to $1$ (every Tweet of the user will be sampled irrespective of the time). We found the value of the parameter when set to $0.44$ (meaning within a span of 3-4 days) led to the maximum validation set accuracy. When the parameter was set at $1$, it caused adding noise, and lost the mapping of the input text to the location label (in this case the zipcode). \looseness = -2

\vspace{0.1cm}
\noindent \textbf{Clustering effect:}
Furthermore, we ran an ablation study to understand the effect of the number of clusters on validation sets accuracy. We trained the model predicting zipcodes with number of clusters ranging between $1$ to $15$, incremented with step size $2$. We found that the accuracy was quite low with \textit{num-cluster} = $1$ (acc: $32.24\%$). It increased to the maximum value with \textit{num-cluster} = $6$ (acc: $65.34\%$). Upon further increasing the \textit{num-clusters} the accuracy dropped to $61.23\%$.
We learned that when there is only one cluster there is no sense of similarity between the Tweets and thus each training input is injected with noise rather than text from similar other data samples. Upon increasing the number of clusters to an optimal value, the training input is concatenated with the encoded representation of similar tokens that map with the location label. However, when the number of clusters is way too many, then there are not adequate number of similar Tweets to look into to inject into the modeling problem. It is noteworthy, to mention that that the clustering based embedding is added only during the training phase of the model. During validation we only use the input Tweet text. \looseness = -2

\vspace{0.1cm}
\noindent \textbf{Neighborhood embedding:}
Finally we were curious to learn about the relevance of the pre-computed neighbor embedding of Tweets based on cosine similarity. When we included the embeddings in the model training process, we achieved a validation accuracy of $71.32\%$ on the neighborhood prediction task. When we trained the model without the embeddings, we found that the validation set accuracy dropped to $58.34\%$. This confirmed that the embeddings' helped the model learn the data better to predict the neighborhood codes. The parameter that controlled the number of similar Tweets to use, was set based on trial and error till we achieved the best performance results for our dataset. In the future, we need further studies to understand the impact of this value to the modeling problem. \looseness = -2

\section{Discussion and Limitation}

\vspace{0.1cm}
\noindent \textbf{User interaction and model exploration:}
The human machine collaborative modeling pipeline presented in the paper, needed spot checking model results and conducting exploratory data analysis to make sense of the results. User interaction was critical in this process, which was deployed using interactive visualizations in Jupyter notebook, leveraging various Python libraries including altair, seaborn, ipyleaflet, matplotlib, and others (see Figure~\ref{fig:diag_flow}). This was adequate for domain expert users who had strong grounding on data analysis and beginner level understanding of ML modeling. However, for domain experts who are not skilled/trained as an analysts, we understand that we need explicit visual interfaces that abstracts user interaction to empower them in the modeling process. With the success of this human machine collaboration in building better performing models for urban planners, we plan to further develop interactive visual interfaces to support this process as the next step.  
Yang et al. confirmed that it is crucial to attain the right level of abstraction in scalable user interaction to employ robust and successful human machine collaborative model development \cite{Yang2019ASO}. In this work, Jupyter notebooks were sufficient to provide the scalable user interaction needed by the urban planner analysts'. However, moving forward we need more further research in abstracting user interactions and scalable interfaces that can seamlessly empower co-developing models as users interact with the model results.
\looseness = -2

\vspace{0.1cm}
\noindent \textbf{User name dictionary:}
Most of the better performing models presented in this paper, relies on the creation of a user name dictionary. Our hypothesis is that, when queried with a new Tweet with username as the meta data, our technique would augment the Tweet text with other text content retrieved from the user name dictionary object. This approach worked very well for us as the exploratory data analysis by domain experts revealed that there are many frequently posting users. However, we are aware that this approach may be  less effective on a dataset on which the user name dictionary is sparse. In that case, we may rely on other meta data content that we have not investigated in this work including hash tags, Tweet mentions, ReTweet counts, and others. In general, we have observed that frequently occurring user names are fairly common and there's a high likelihood of creating a dense user name dictionary that helps improve the models' prediction. \looseness = -2

\vspace{0.1cm}
\noindent \textbf{Fine-grained prediction:}
In this work, urban planners were motivated to build models for fine-grained Tweet prediction. In that endeavor, we started with latitude and longitude prediction. While we observed some success in their prediction, given the noise and ambiguity in short text posts in Twitter, improving accuracy closer to the Tweet posts less than 30 miles (as desired by end-users) seemed a bit too ambitious. Courtesy to the collaboration with the domain experts, we were able to adjust our analytical goal from longitudes to zipcode, and then to neighborhood level prediction. Noteworthy to mention that both zipcodes and neighborhood level predictions are still fine-grained compared to what we observed from the literature that most of the previous work in geolocation prediction have been at the level of country or city. In the future, we plan to continue our current research in predicting fine-grained Tweet geolocation based on the lessons learned from the conducted experiments. \looseness = -2
% build upon the less

%% file: content/conclusion.tex
\section{Conclusion}

In this paper, we demonstrate our investigation into modeling fine-grained geolocation prediction of non geo-tagged Tweets at multiple levels of granularity including zipcodes, neighborhoods, and longitude with latitude information. In the modeling process, we include humans (end-users) to directly adjust model architecture providing an opportunity to feed their domain expertise to better solve their analytical problem. We collaborate with urban planners, in which we support them geolocate Tweets to better analyse peoples' sentiment as expressed through Twitter posts.
Through a set of experiments on a variety of deep neural network architectures, we show our data analysis pipeline supports training a robust model that generalizes well to unseen Tweet posts. \looseness = -2
\looseness = -2